\documentclass[conference]{IEEEtran}
\IEEEoverridecommandlockouts
\usepackage{cite}
\usepackage{amsmath,amssymb,amsfonts}
\usepackage{utfsym}
\usepackage{algorithmic}
\usepackage{graphicx}
\usepackage{textcomp}
\usepackage{xcolor}
\usepackage{multirow}
\def\BibTeX{{\rm B\kern-.05em{\sc i\kern-.025em b}\kern-.08em
    T\kern-.1667em\lower.7ex\hbox{E}\kern-.125emX}}
\begin{document}

\title{Semantic-Aware Mixup for Domain Generalization\\
{}
}

\author{\IEEEauthorblockN{Chengchao Xu\IEEEauthorrefmark{1}, 
Xinmei Tian\IEEEauthorrefmark{1}\IEEEauthorrefmark{2}\thanks{Corresponding author: Xinmei Tian (email: xinmei@ustc.edu.cn).}}
\IEEEauthorblockA{\IEEEauthorrefmark{1}University of Science and Technology of China, Hefei, China}
\IEEEauthorblockA{\IEEEauthorrefmark{2}Institute of Artificial Intelligence, Hefei Comprehensive National Science Center, Hefei, China}
}

\maketitle

\begin{abstract}
 Deep neural networks (DNNs) have shown exciting performance in various tasks, yet suffer generalization failures when meeting unknown target domains. One of the most promising approaches to achieve domain generalization (DG) is generating unseen data, e.g., mixup, to cover the unknown target data. However, existing works overlook the challenges induced by the simultaneous appearance of changes in both the semantic and distribution space. Accordingly, such a challenge makes source distributions hard to fit for DNNs. To mitigate the hard-fitting issue, we propose to perform a \emph{s}emantic-\emph{a}ware \emph{m}ixup (SAM) for domain generalization, where whether to perform mixup depends on the semantic and domain information. The feasibility of SAM shares the same spirits with the Fourier-based mixup. Namely, the Fourier phase spectrum is expected to contain semantics information (relating to labels), while the Fourier amplitude retains other information (relating to style information). Built upon the insight, SAM applies different mixup strategies to  the Fourier phase spectrum and amplitude information. For instance, SAM merely performs mixup on the amplitude spectrum when both the semantic and domain information changes. Consequently, the overwhelmingly large change can be avoided. We validate the effectiveness of SAM using image classification tasks on several DG benchmarks.
\end{abstract}


\section{Introduction}
The past few years have witnessed the remarkable performance of deep learning in diverse application areas such as image understanding~\cite{AlexKrizhevsky2012ImageNetCW, JonathanTompson2014JointTO}, speech recognition~\cite{TomasMikolov2011StrategiesFT, GeoffreyEHinton2012DeepNN}, and natural language processing~\cite{IlyaSutskever2014SequenceTS, RuhiSarikaya2014ApplicationOD}. Among these exciting achievements, deep neural networks (DNNs) make a significant contribution~\cite{IanGoodfellow2016DeepL, YannLeCun2015DeepL, KaimingHe2015DeepRL}. However, even advanced DNNs perform poorly on unknown target domains due to the distribution discrepancy between source and target domains~\cite{DaLi2017DeeperBA}. In many practical scenarios, such distribution discrepancies between source and target domains are unavoidable, attracting attention to domain generalization~\cite{RohanTaori2020MeasuringRT, ShaiBenDavid2010ATO, BenjaminRecht2019DoIC, JoseGMorenoTorres2012AUV}.

A possible approach to reduce the negative influence caused by the distribution discrepancy is domain adaptation (DA) which establishes a connection between the source domain(s) and a specific target domain with limited available data~\cite{wang2018deep}. However, DA relies on a relatively strong assumption that DNNs have access to the (labeled or unlabeled) target data, which could be violated in some practical scenarios. To release the assumption, domain generalization (DG) is proposed~\cite{DaLi2017DeeperBA, muandet2013domain}, where DNNs merely have access to source data. Specifically, DG aims to learn a model using data from multiple semantically related source domains and evaluates the trained DNNs using unseen target domains. This is different from the scenarios considered in OOD detection~\cite{fangout}. Many existing methods have shown encouraging results utilizing adversarial training\cite{li2018deep, li2018domain, shao2019multi}, meta-learning\cite{DaLi2017LearningTG, YogeshBalaji2018MetaRegTD, li2019feature, dou2019domain}, self-supervised learning\cite{FabioMariaCarlucci2019DomainGB}, or domain augmentation\cite{volpi2018generalizing, shankar2018generalizing, zhou2020deep, zhou2020learning} techniques to tackle generalization failures.

\begin{figure}[tbp]
\centering
\includegraphics[width=0.8\linewidth]{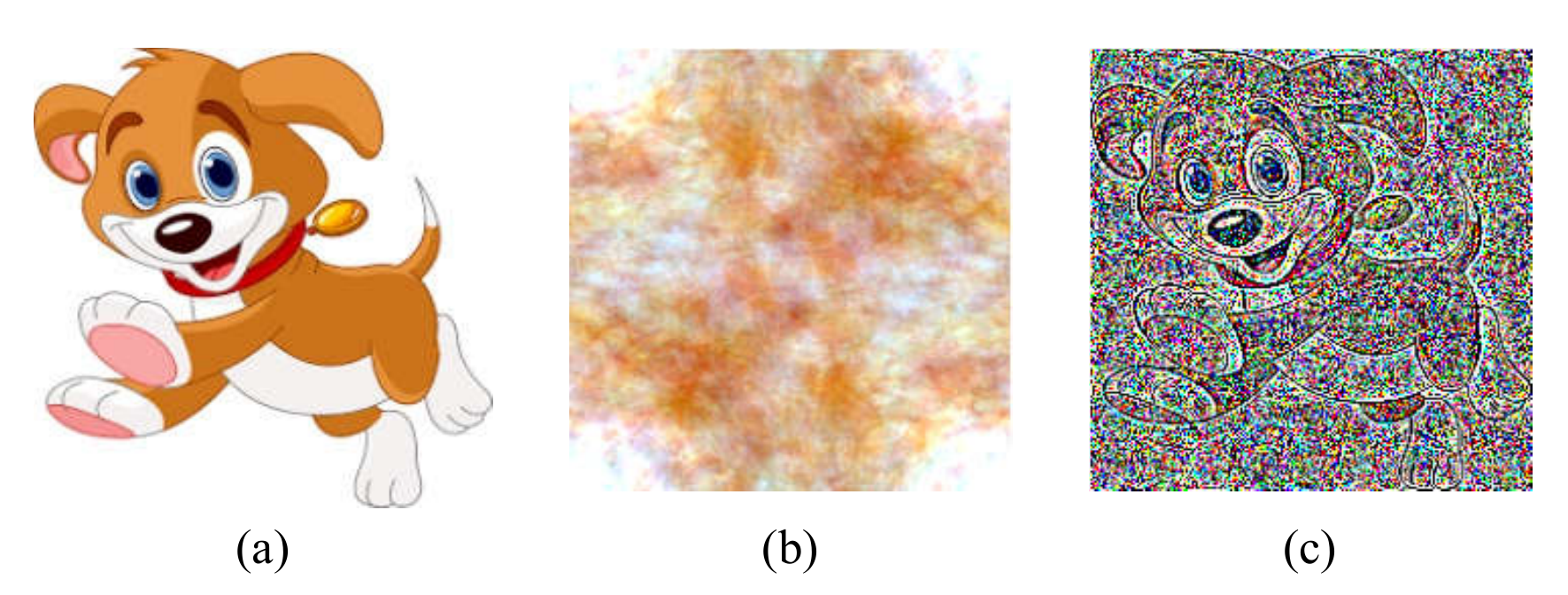}
\caption{Examples of amplitude-only and phase-only reconstruction: (a) original image; (b) reconstructed image with amplitude information by setting the phase component to a constant; (c) reconstructed image with phase information by setting the amplitude component to a constant.}
\label{fig2}
\end{figure}

Among these insightful approaches, domain generalization with Fourier-based assumption has shown great potential in tackling generalization failures~\cite{xu2021fourier}. The insight of \cite{xu2021fourier} is the same as previous fruitful studies~\cite{AlanVOppenheim1979PhaseIS, AlanVOppenheim1981TheIO, BruceCHansen2007StructuralSA, LeonNPiotrowski1982ADO}. These studies show that the phase component in the Fourier spectrum of signals includes the most high-level semantics of the original signals. In contrast, the amplitude component in the Fourier spectrum mainly retains low-level statistics, e.g., style information. The corresponding vivid illustration is provided in Fig.\ref{fig2}, where (Fourier) reconstructing an image Fig.\ref{fig2} (a) using phase information Fig.\ref{fig2} (c) provides sufficient information for predicting the label, i.e., a dog. Meanwhile, (Fourier) reconstruction using amplitude information Fig.\ref{fig2} (b) seems related to the style information of the original image. Namely, in Fig.\ref{fig2}(b), it is relatively hard to recognize the exact object from the amplitude-only reconstruction, namely, the amplitude component mainly retains low-level information related to style information. On the contrary, thanks to the high-level semantics, the phase-only reconstruction demonstrates the important visual structures. Specifically, one can relatively easily recognize the "dog" conveyed in the original image, as shown in Fig.\ref{fig2}(c).

Built upon the Fourier assumption, advanced works have achieved exciting performance on domain generalization~\cite{yang2020fda,xu2021fourier}. Specifically, \cite{yang2020fda} proposes a simple yet effective image translation strategy to endow DNNs with robustness against distribution discrepancy, where the amplitude spectrum of a source image is replaced using a randomly selected target image. \cite{xu2021fourier} develops a Fourier-based mixup, which linearly interpolates between the amplitude spectrums of two images. Namely, \cite{xu2021fourier} proposed to perform mixup \cite{HongyiZhang2017mixupBE} in the amplitude space, outperforming existing DG methods.

 \begin{figure*}[htbp]
\centering
\includegraphics[width=0.8\linewidth]{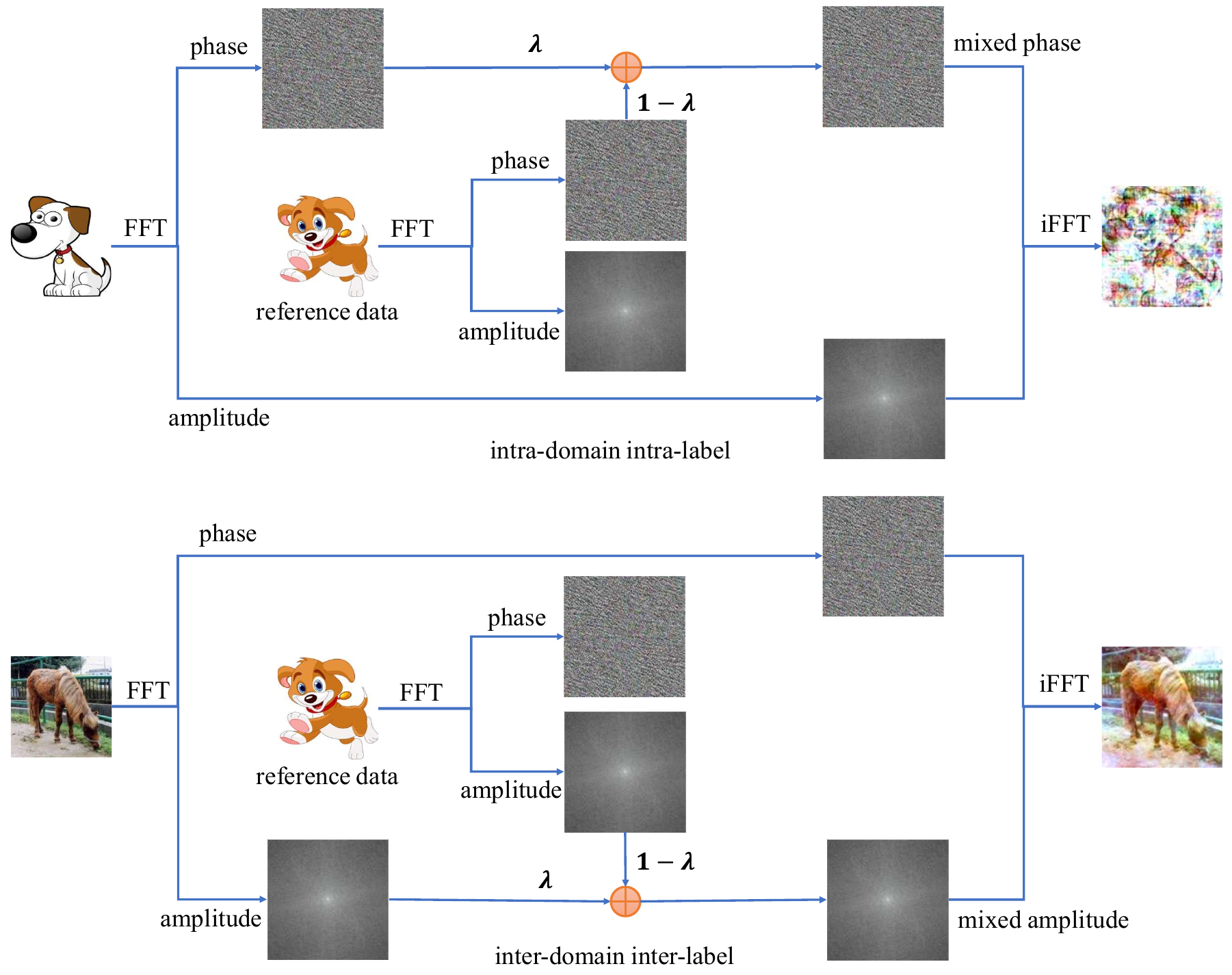}
\caption{Illustration of SAM. Fourier transformation is applied to each image, leading to the phase spectrum and the amplitude spectrum. Which spectrum (phase or amplitude) to perform mixup depends on the label and domain information.}
\label{fig1}
\end{figure*}

However, existing Fourier-based data augmentation approaches overlook the challenges induced by the simultaneous appearance of changes in both the semantic and distribution space. For instance, these methods perform mixup when both the semantic and domain information are significantly changed. Consequently, training models using hard data can cause performance degradation~\cite{pang2018towards}. Namely, such a challenge makes source data hard to fit. Accordingly, training models with such a rough version of mixup leads to adverse impacts on achieving domain generalization.

To tackle the hard-to-fit issue, we propose to perform a semantic-aware mixup (SAM) for domain generalization, inspired by the content and style decomposition of data~\cite{causal}. In SAM, whether and how to perform data augmentation depends on the semantic and domain information. Specifically, SAM realizes data augmentation using mixup and performs mixup using both the label (semantic) and domain information, such that the overwhelmingly large change can be avoided. The proposed Fourier-based data augmentation method shares the same inspiration with~\cite{xu2021fourier}. More specifically, in SAM, whether to perform mixup depends on the semantic (label) and style (domain) information of images, where the phase (amplitude) spectrum of Fourier transformation stands for the semantic (style) information. For Fourier spectrums, not only the amplitude spectrums but also the phase ones make a difference when doing mixup. 

A vivid illustration of SAM is given in Fig. \ref{fig1}, where the phase (amplitude) spectrum represents the semantic (style) information. The intuition is straightforward. The phase spectrum implies high-level semantics. Thus, it is necessary to perform interpolation in the phase spectrum during data augmentation, making the classification boundary clear. Similarly, the amplitude spectrum represents a low-level style, thus, it is also necessary to perform interpolation in the amplitude spectrum during data augmentation, making DNNs robust against distribution shifts. Note that, it is needless to perform mixup for samples drawing from the same distribution (domain), since these samples have similar amplitude spectrums. Moreover, if both the semantics and style are different from the two samples, the SAM will perform mixup merely in the amplitude spectrum. This is because the simultaneous appearance of changes in both the semantic and distribution space makes the augmented data hard to fit for DNNs. Finally, to ensure the consistency of model prediction, we introduce an explicit constraint that encourages DNNs to make consistent predictions on original and augmented images.

We perform empirical verification to support the insights and the effectiveness of SAM. Specifically, we evaluate SAM on three widely used datasets PACS\cite{DaLi2017DeeperBA}, OfficeHome\cite{HemanthVenkateswara2017DeepHN}, and Digits-DG\cite{zhou2020deep}. The experimental results demonstrate that our approach outperforms state-of-the-art DG methods. Moreover, we conduct an ablation study for an in-depth analysis of SAM, providing more empirical insights.

\section{Related Work}
\subsection{Domain Generalization}
Domain generalization (DG) \cite{zhou2022domain} aims to improve model performance in scenarios where the source and target domain distributions are statistically different. It is similar to domain adaptation (DA) \cite{wang2018deep} where the domain gap also exists. However, DG assumes the (labeled or unlabeled) target data samples are inaccessible, while DA supposes such data are available during training. Thus, DG aims to extract representations from multiple source domains to generalize well on unseen target domains.

The past few years have witnessed promising improvements for DG. One of the most effective DG approaches focuses on distribution alignment, aiming to minimize the discrepancy between source and target domains. These methods can be roughly divided into two categories: kernel methods\cite{muandet2013domain, ghifary2016scatter} and domain-adversarial learning\cite{li2018deep, li2018domain, shao2019multi, jia2020single}. 
To capture generalizable features, \cite{li2018learning} proposes a meta-learning strategy to model the meta-features for DG, sharing the same spirit with the training strategy of MAML~\cite{finn2017model}. These methods divide source data into meta-train and meta-test sets. Accordingly, these methods train models using the meta-train set, hoping to improve the performance on the meta-test set. The meta-learning insight is also employed for learning a regularizer~\cite{YogeshBalaji2018MetaRegTD}, a feature-critic network\cite{li2019feature}, or how to maintain semantic relationships\cite{dou2019domain}. 

Domain augmentation\cite{volpi2018generalizing, shankar2018generalizing, zhou2020deep, zhou2020learning} has shown promising potential to solve problems of DG failures by generating unseen samples. Recently, \cite{shi2020informative} biases their model to shape features by filtering out texture features according to local self-information, considering the robustness of a shape-biased model to out-of-distributions\cite{geirhos2018imagenet}. Similarly, \cite{FabioMariaCarlucci2019DomainGB} addresses a self-supervised jigsaw task to help the model learn the global shape features. Built upon these works, \cite{wang2020learning} incorporates a momentum metric learning scheme. Advanced works develop a Fourier-based data augmentation by emphasizing the Fourier phase information~\cite{xu2021fourier}, achieving state-of-the-art performance. In contrast, our method proposes to perform mixup using the semantic and domain information.

\subsection{Information of Fourier spectrums}
Previous insightful studies\cite{AlanVOppenheim1979PhaseIS, AlanVOppenheim1981TheIO, BruceCHansen2007StructuralSA, LeonNPiotrowski1982ADO} have shown that the phase spectrum in the Fourier transformation retains most of the high-level semantics in the original signals, while the amplitude spectrum mainly contains low-level statistics. Built upon these observations, \cite{yang2020fda} applies the Fourier-based perspective to domain adaptation. It reduces the discrepancy between the source and target distributions by swapping the low-frequency spectrum with the others. An advanced work \cite{yang2020phase} introduces the phase consistent under domain adaptation, which performs better than the commonly used cycle consistency~\cite{hoffman2018cycada} for segmentation. In contrast, we consider not only the phase spectrums but also the amplitude ones for domain generalization, involving both the semantic and domain information.

\subsection{Consistency regularization}
Consistency regularization (CR) is widely used to reduce overfitting in supervised and semi-supervised learning~\cite{laine2016temporal,tarvainen2017mean}. \cite{laine2016temporal} first proposes a consistency loss for classification tasks. \cite{tarvainen2017mean} further uses a momentum-updated teacher to provide better targets for consistency alignment. Some methods \cite{miyato2018virtual, park2018adversarial} replace the stochastic perturbations with adversarial ones. \cite{verma2019interpolation} shows that a interpolation consistency with mixup\cite{HongyiZhang2017mixupBE} can benefit generalization performance. Besides, \cite{french2017self, shu2018dirt, wu2020dual} also use consistency regularization to endow DNNs better generalizability on target domains. Similar to these works, we employ a consistency constraint to encourage DNNs to make consistent predictions on original and augmented images.


\section{Method}
In this section, we will introduce preliminaries for domain generalization, and give details of semantic-aware mixup (SAM) proposed to tackle generalization failures.  

\subsection{Preliminaries for domain generalization}
In DG, DNNs are typically assumed to have the access to a training set (sampled from source domains) composed of $S$ source domains $\mathcal{D} = \{\mathcal{D}_1, \mathcal{D}_2, \dots, \mathcal{D}_S\}$ with $N_k$ labeled samples $\{(x_i^k,y_i^k)\}_{i=1}^{N_k}$ in the $k$-th domain $\mathcal{D}_k$, where $x_i^k$ and $y_i^k \in \{1, ..., C\}$ denote the inputs and labels respectively. For classification tasks, domain generalization aims to learn a model $f(\cdot;\theta)$ to predict the label $\hat{y}_i^k$ corresponding to the input $x_i^k$ with a good performance on an unseen domain $\mathcal{D}_t$.
For simplicity, we encode label $y_i^k$ into the one-hot label $z_i^k\in\mathbb{R}^C$, following \cite{HongyiZhang2017mixupBE}, where the $y_i^k$-th component is one and zeroes for the others.

\subsection{Fourier Transformation Assumption}
Following \cite{xu2021fourier}, we regard Fourier transformation as a competitive candidate to represent semantic and non-semantic information. Thus, it is necessary to revisit the formulation of Fourier transformation before giving further details:
\begin{equation*}
    \mathcal{F}(x)(u,v) = 
    \sum_{h=0}^{H-1}\sum_{w=0}^{W-1}
    x(h,w)e^{-j2\pi \left(\frac{h}{H}u+\frac{w}{W}v\right)},\label{eq1}
\end{equation*}
where $x$ is a single-channel image, $W$ ($H$) represents width (height) of $x$, and $u$ ($v$) stands for the index used in Fourier transformation. The amplitude spectrum $\mathcal{A}(x)$ and the phase spectrum $\mathcal{P}(x)$ can be formulated as follows:
\begin{equation*}
\begin{split}
    \mathcal{A}(x)(u, v) &= \left[R^2(x)(u, v) + I^2(x)(u, v)\right]^\frac{1}{2},
    \\
    \mathcal{P}(x)(u, v) &= \arctan\left[\frac{I(x)(u, v)}{R(x)(u, v)}\right],
    \label{eq3}
\end{split}
\end{equation*}

where $R(x)$ and $I(x)$ denote the real and imaginary part of Fourier transformation $\mathcal{F}(x)$, respectively. Accordingly, the inverse Fourier transformation $\mathcal{F}^{-1}(\mathcal{A}(x), \mathcal{P}(x))$ is defined as:
\begin{align*}
    &\mathcal{F}^{-1}
    (\mathcal{A}(x)(u, v), \mathcal{P}(x)(u, v))
    (h, w) \\
    &= \sum_{u=0}^{H-1}\sum_{v=0}^{W-1}\mathcal{A}(x)(u, v)e^{j\left[2\pi \left(\frac{h}{H}u+\frac{w}{W}v\right)+\mathcal{P}(x)(u, v)\right]}.
\end{align*}
When dealing with RGB images, Fourier transformation is typically calculated independently for each channel. Fourier transformation and inverse Fourier transformation can be calculated using the FFT algorithm \cite{angel1982fast} for efficiency.

According to previous work\cite{AlanVOppenheim1979PhaseIS, AlanVOppenheim1981TheIO, BruceCHansen2007StructuralSA, LeonNPiotrowski1982ADO}, the Fourier phase spectrum $\mathcal{P}(x)$ is expected to contain semantics information, while the Fourier amplitude $\mathcal{A}(x)$ spectrum retains other information. Specifically, the phase component of the Fourier spectrum preserves the high-level semantics of the original signal, while the amplitude component contains low-level statistics. Built upon the assumption, advanced works have achieved promising performance by performing mixup merely in the amplitude space:
\begin{equation} \label{fmix1}
\begin{split}
    & \tilde{y} = 
    \lambda y + (1-\lambda)y_r
    , \\
    &
    \tilde{x} = 
    \mathcal{F}^{-1}
    (\mathcal{\tilde{A}}(x), \mathcal{P}(x)).
\end{split}
\end{equation}
where $\lambda \sim U(0,\eta)$ represents how to perform mixup between two samples, the hyperparameter $\eta$ controls the strength of mixup, and the modified amplitude $\mathcal{\tilde{A}}(x)$ of $x$ is calculated using a sample $x_r$ with its label $y_r$ randomly selected for performing mixup:
\begin{equation} \label{fmix2}
    \mathcal{\tilde{A}}(x) = 
    \lambda \mathcal{A}(x) + (1-\lambda)\mathcal{A}(x_r).
\end{equation}
According to the Fourier assumption, augmenting data using Eq.\eqref{fmix1} and Eq.\eqref{fmix2} is equal to performing mixup in the style space, making DNNs trained on these samples robust against distribution shifts.

\subsection{Semantic-Aware Mixup}


Inspired by the Fourier assumption, we propose to perform a semantic-aware mixup, SAM, to achieve domain generalization. Specifically, SAM splits the relationship of any two images into four categories, according to the domain and label information. As shown in Table. \ref{tab2}, for two samples $(x_i^k, y_i^k)$ with domain $k$ and $(x_{j}^{l},y_{j}^{l})$ with domain $l$, we define their pair relation as follows:
\begin{enumerate}
    \item \label{case1} intra-domain and intra-label if 
    $$k=l \text{ and } y_i^k = y_{j}^{l}$$
    \item \label{case2} intra-domain and inter-label if 
    $$k=l \text{ and } y_i^k \neq y_{j}^{l}$$
    \item \label{case3} inter-domain and intra-label if 
    $$k\neq l \text{ and } y_i^k = y_{j}^{l}$$
    \item \label{case4} inter-domain and inter-label if 
    $$k\neq l \text{ and } y_i^k \neq y_{j}^{l}.$$
\end{enumerate}

    In \ref{case1}) and \ref{case2}), it is no need to interpolate the amplitude spectrums linearly. On the contrary, interpolation on the phase spectrums can make the model more generalizable to high-level semantics of images:
\begin{equation} \label{fmix3}
\begin{split}
    & \tilde{y} = 
    \lambda y^k_i + (1-\lambda)y^l_j
    , \\
    &
    \tilde{x} = 
    \mathcal{F}^{-1}
    (\mathcal{A}(x^k_i), \mathcal{\tilde{P}}(x^k_i)),
\end{split}
\end{equation}
where the phase spectrum is calculated as follows:
    \begin{equation}\label{eq4}
        \mathcal{\tilde{P}}(x_i^k) = (1-\lambda)\mathcal{P}(x_i^k) + \lambda \mathcal{P}(x_{j}^{l}),
    \end{equation}
    where $\lambda \sim U(0,\eta)$, and hyperparameter $\eta$ controls the strength of the augmentation. 
    
     Compared with \ref{case1}) and \ref{case2}), the domains of the two images in \ref{case3}) are different. We further add interpolation on the amplitude spectrums so as to learn more information in styles in \ref{case3}): 
\begin{equation} \label{fmix4}
\begin{split}
    & \tilde{y} = 
    \lambda y^k_i + (1-\lambda)y^l_j
    , \\
    &
    \tilde{x} = 
    \mathcal{F}^{-1}
    (\mathcal{\tilde{A}}(x^k_i), \mathcal{\tilde{P}}(x^k_i)),
\end{split}
\end{equation}
where the phase spectrum $\mathcal{\tilde{P}}(x_i^k)$ is the same as Eq. \eqref{fmix4} and the amplitude spectrum $\mathcal{\tilde{A}}(x_i^k)$ is calculated as follows:
     \begin{equation}\label{eqa}
        \mathcal{\tilde{A}}(x_i^k) = (1-\lambda)\mathcal{A}(x_i^k) + \lambda \mathcal{A}(x_{j}^{l}),
    \end{equation}

    In \ref{case4}), since the difference in both domains and labels, it is difficult for the classification model to learn from the augmented samples produced by interpolation on both the amplitude and the phase spectrums. Therefore, considering the phase information have better generalization ability across domains, we remove the interpolation on the phase spectrums and use original labels as the augmented ones:
    \begin{equation} \label{fmix4}
\begin{split}
    & \tilde{y} = 
    \lambda y^k_i + (1-\lambda)y^l_j
    , \\
    &
    \tilde{x} = 
    \mathcal{F}^{-1}
    (\mathcal{\tilde{A}}(x^k_i), \mathcal{P}(x^k_i)),
\end{split}
\end{equation}
where $\mathcal{\tilde{A}}(x^k_i)$ can be calculated using Eq. \eqref{eqa}.

Table \ref{tab2} summarizes the strategy mentioned above, and Fig. \ref{fig1} shows the procedure of the modified Fourier augmentation.

    \begin{table}[htbp]
    \caption{The mixup strategy of SAM.}
    \begin{center}
    \begin{tabular}{c|c|c|c}
    \hline
     Domains  &  Labels & Mixup on amplitude& Mixup on phase  \\
    \hline
    \textbf{intra}-domain & \textbf{intra}-label & $\usym{2717}$ & $\usym{2713}$ \\
    \textbf{intra}-domain & \textbf{inter}-label & $\usym{2717}$ & $\usym{2713}$ \\
    \textbf{inter}-domain & \textbf{intra}-label & $\usym{2713}$ & $\usym{2713}$ \\
    \textbf{inter}-domain & \textbf{inter}-label & $\usym{2713}$ & $\usym{2717}$ \\
    \hline
    \end{tabular}
    \label{tab2}
    \end{center}
    \end{table}

After applying SAM, the augmented data are fed to DNNs, where the objective function can be realized as follows:
    \begin{equation}\label{eq10}
        \mathcal{L}(\theta) = 
        \mathbb{E}_{x,y}\left[
        \ell(f(x;\theta), y)
        +
        \gamma
        \ell(f(\tilde{x};\theta), \tilde{y})\right],
    \end{equation}
where $\theta$ is the model parameter, $f(\cdot;\theta)$ is the model output, e.g., probability vector, $\ell$ stands for the loss function, e.g., cross-entropy loss, $(\tilde{x}, \tilde{y})$ represents the augmented input pair generated using $(x, y)$, and $\gamma$ is a hyper-parameter trading off the original and augmented data.

\subsection{Consistency Constraint}
Recall that the generated data are typically different from the source data, resulting in different distributions. Thus, if the generated data are drastically different from the source data, DNNs trained over generated data would overfit the generated ones. Consequently, the performance of DNNs trained over generated data would be degraded. To avoid the model shift issue, we further propose a consistency constraint, where the model outputs generated with the current and past model parameters are consistent. Inspired by MoCo~\cite{he2020momentum}, we use a momentum-updated teacher model to provide anchors for consistency alignment as suggested in \cite{AnttiTarvainen2017MeanTA}. Specifically, using an exponential moving average, the teacher model $\theta_t$ receives parameters from the current model $\theta$ during training:
\begin{equation}
    \theta_{t} = m\theta_{t} + (1-m)\theta,
\end{equation}
where $m$ stands for the momentum hyper-parameter.
To measure the consistency, we employ Kullback-Leibler divergence:
\begin{align}
    \mathcal{L}_{c1} &= \text{KL}
    (f(\tilde{x}_i^k;\theta) ||
    f(x_i^k;\theta_t)), \\
    \mathcal{L}_{c2} &= \text{KL}
    (f(x_i^k;\theta) ||
    f(\tilde{x}_i^k;\theta_t)).
\end{align}

Finally, the overall objective function of SAM is $\mathcal{L}_{all}$:
\begin{equation}
\min_{\theta}
    \mathcal{L}_{all}(\theta) = \mathcal{L}(\theta) 
    +
    \beta (
    \mathcal{L}_{c1}(\theta) + \mathcal{L}_{c2}(\theta)
    ),
\end{equation}
where the hyper-parameter $\beta$ balances the classification loss and the consistency constraint.

\section{Experiment}
In this section, we evaluate SAM on benchmark datasets for multi-domain image classification. We also provide ablation studies to show some insights into SAM. The code to reproduce our results is available~\footnote{https://github.com/ccxu-ustc/SAM}.

\subsection{Experimental setup}
To evaluate the effectiveness of our method, following \cite{xu2021fourier}, we use three datasets: PACS\cite{DaLi2017DeeperBA}, OfficeHome\cite{HemanthVenkateswara2017DeepHN} and Digits-DG\cite{zhou2020deep}. PACS includes 9,991 images of 4 domains: Art-painting (A), Cartoon (C), Photo (P), and Sketch (S). It contains seven same labels for each domain, and there exists a large discrepancy in image styles between different domains. We use the original train-validation split provided by \cite{DaLi2017DeeperBA}. OfficeHome contains 15,500 images of 65 classes from four domains (Art, Clipart, Product, and Real-World). We randomly split each domain into 90\% for training and 10\% for validation. Digits-DG is a digit recognition benchmark including four datasets: MNIST\cite{YannLeCun1998GradientbasedLA}, MNIST-M\cite{YaroslavGanin2015UnsupervisedDA}, SVHN\cite{YuvalNetzer2011ReadingDI}, and SYN\cite{YaroslavGanin2015UnsupervisedDA}. We use the original train-validation split in \cite{zhou2020deep}.

We use the widely used strategy, i.e., leave-one-domain-out, for evaluation. For the training process, we train our model on the training splits and select the best model on the validation splits of all source domains. For the testing process, we evaluate the selected model on all images of the held-out domain. We run every result three times independently and report the average classification accuracy.

\subsection{Implementation details}
We generally follow the implementations of \cite{FabioMariaCarlucci2019DomainGB, zhou2020deep, xu2021fourier}. 
For all experiments, the network is trained by using Nesterov-momentum SGD with a momentum of 0.9 and weight decay of 5e-4 for 50 epochs. We set momentum $m$ for the teacher model to 0.9995. We also use a sigmoid rampup~\cite{AnttiTarvainen2017MeanTA} for $\beta$ with a length of 5 epochs. The hyper-parameter $\gamma$ is set to $1.0$.

For PACS and OfficeHome, following our baselines, we use the ImageNet~\cite{JiaDeng2009ImageNetAL} pre-trained ResNet\cite{KaimingHe2015DeepRL} as our backbone. The batch size is 16, and the initial learning rate is 0.001 and decayed by 0.1 at 80\% of the total epochs. The weight $\beta$ of the consistency loss is chosen as 2 for PACS and 200 for OfficeHome.  The augmentation strength $\eta$ is set to 1.0 for PACS and 0.2 for OfficeHome.
For Digits-DG, We use the same backbone network as \cite{zhou2020deep, zhou2020learning}.  The batch size is 128, and the initial learning rate is 0.05 and decayed by 0.1 every 20 epochs. The weight $\beta$ of the consistency loss and the augmentation strength $\eta$ is chosen as 2 and 1.0, respectively.

Note that the size of the batch fed into the models is actually $4\times$ batch size because two original images together with two augmented images are generated by an original sample during the modified Fourier data augmentation.

Following the previous protocol \cite{FabioMariaCarlucci2019DomainGB}, for PACS and OfficeHome, we randomly crop the images to retain between 80\% to 100\%. We also apply random horizontal flipping and random color jittering with a magnitude of 0.4. The input size is $224\times 224$. For Digits-DG, the input size is $32\times 32$.

\subsection{Evaluation on PACS}
\begin{table}[htbp]
\caption{Model accuracy of leave-one-domain-out evaluation on PACS. The best and second-best results are bolded and underlined, respectively.}
\begin{center}
\resizebox{\linewidth}{!}{
\begin{tabular}{l|cccc|c}
\hline
Methods          & Art            & Cartoon        & Photo          & Sketch         & Avg.           \\ \hline
DeepAll          & 84.94          & 76.98          & \underline{97.64} & 76.75          & 84.08          \\
MetaReg\cite{YogeshBalaji2018MetaRegTD}          & 87.20          & 79.20          & 97.60          & 70.30          & 83.60          \\
MASF\cite{QiDou2019DomainGV}             & 82.89          & 80.49          & 95.01          & 72.29          & 83.67          \\
EISNet\cite{ShujunWang2020LearningFE}           & 86.64          & 81.53          & 97.11          & 78.07          & 85.84          \\
RSC\cite{ZeyiHuang2020SelfchallengingIC}              & 87.89          & \underline{82.16}          & \textbf{97.92}          & 83.35          & 87.83          \\
FACT\cite{xu2021fourier} & \textbf{89.68} & 81.06          & 97.02          & \underline{83.76}    & \underline{87.88}    \\ 
\textbf{Ours}             & \underline{88.85}$\pm$0.66    & \textbf{82.30}$\pm$0.57 & 97.21$\pm$0.18   & \textbf{85.83}$\pm$0.21 & \textbf{88.55}  \\ \hline
\end{tabular}
}
\label{tab1}
\end{center}
\end{table}


\begin{figure*}[htbp]
    \centering
    \includegraphics[width=\linewidth]{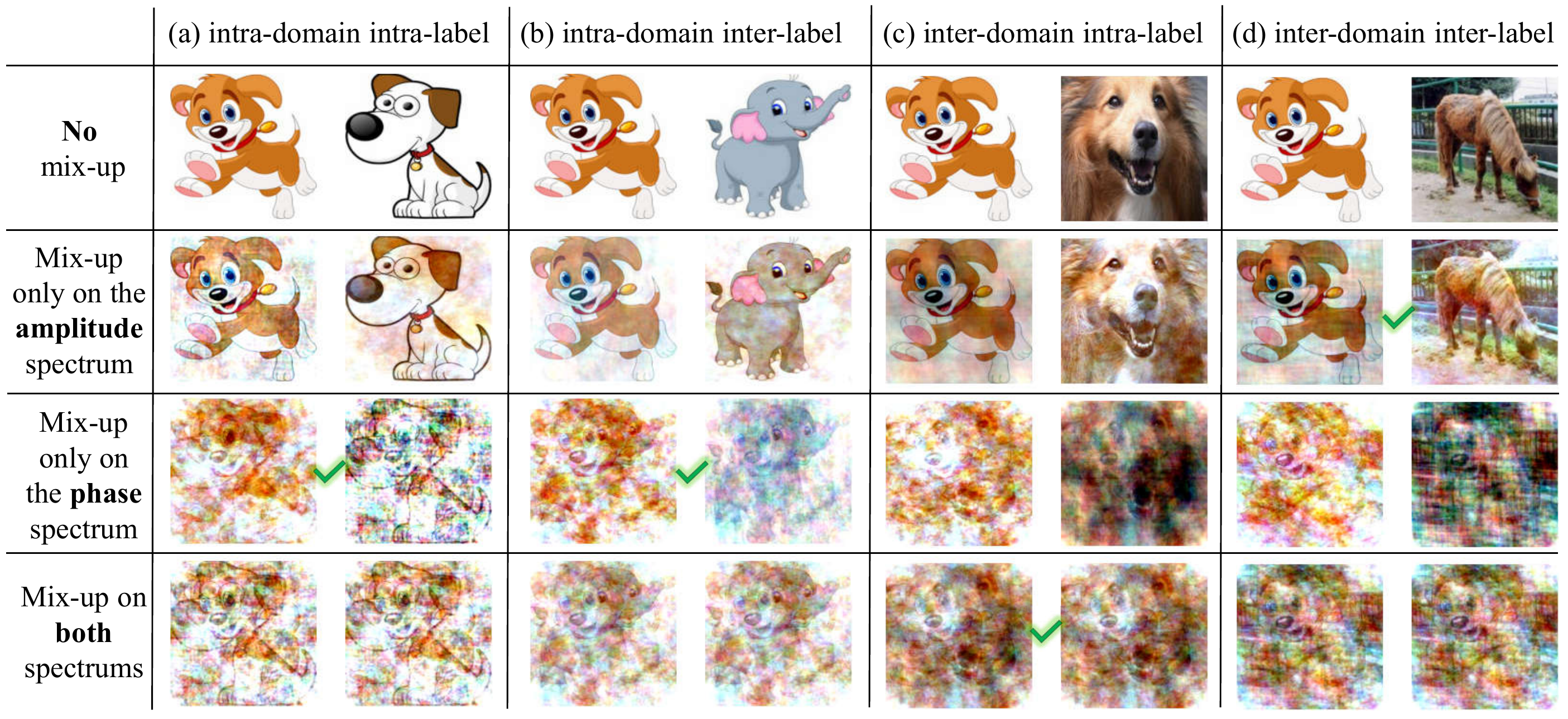}
    \caption{Visualization of four kinds of pair images after different kinds of mixup strategy. The results of SAM are labeled with a green check mark.}
    \label{fig4}
\end{figure*}

The results are shown in Table \ref{tab1}, where DeepAll is the baseline that employs a na\"ive ConvNet trained from all source data. Our method achieves the best performance among the top-performing ones. Note that the DeepAll baseline can get excellent accuracy on the photo domain in which the images are similar to ones in the pretrained dataset ImageNet. However, DeepAll performs badly on the art-painting, cartoon, and sketch domains due to a large domain gap. Our approach is 3.91\%, 5.32\%, and 9.08\% ahead of the baseline on the art-painting, cartoon, and sketch, respectively. For the photo domain, with the Fourier-based data augmentation mentioned above, our method pays more attention to high-level information to improve the generalization ability compared to Baseline. Therefore, some sacrifice on low-level cues may lead to a performance drop since the photo domain contains lots of intricate details. Nevertheless, our approach still gets a better performance overall.

Compared with SOTA, Our approach performs better than FACT\cite{xu2021fourier} because of the changes dealing with the situation for domains and labels in detail. Besides, our method also outperforms MASF\cite{QiDou2019DomainGV}, which is based on meta-learning. In sum, these experimental results show the effectiveness of the proposed method.

\subsection{Evaluation on OfficeHome}

\begin{table}[htbp]
\caption{Model accuracy of leave-one-domain-out evaluation on OfficeHome. The best and second-best results are bolded and underlined, respectively.}
\begin{center}
\resizebox{\linewidth}{!}{
\begin{tabular}{l|cccc|c}
\hline
Methods          & Art            & Clipart        & Product          & Real         & Avg.           \\ \hline
DeepAll          & 57.88          & 52.72          & 73.50 & 74.80          & 64.72          \\
CCSA\cite{motiian2017unified}            & 59.90          & 49.90         & 74.10          & 75.70          & 64.90          \\
MMD-AAE\cite{li2018domain}             & 56.50          & 47.30          & 72.10          & 74.80          & 62.70          \\
CrossGrad\cite{shankar2018generalizing}           & 58.40          & 49.40          & 73.90          & 75.80          & 64.40          \\
RSC\cite{ZeyiHuang2020SelfchallengingIC}              & 58.42          & 47.90          & 71.63          & 74.54          & 63.12          \\
DDAIG\cite{zhou2020deep}           & 59.20          & 52.30          & \underline{74.60}          & 76.00          & 65.50          \\
Jigen\cite{FabioMariaCarlucci2019DomainGB}           & 53.04          & 47.51          & 71.47          & 72.79          & 61.20          \\
L2A-OT\cite{zhou2020learning}           & \underline{60.60}          & 50.10          & \textbf{74.80}          & \textbf{77.00}         & 65.60          \\
FACT\cite{xu2021fourier}             & 60.34         & \underline{54.85}         & 74.48          & \underline{76.55} & \underline{66.56}          \\
\textbf{Ours}     & \textbf{60.79}$\pm$0.58    & \textbf{55.47}$\pm$0.10 & 74.37$\pm$0.12    & 76.37$\pm$0.10 & \textbf{66.75} \\ \hline
\end{tabular}
}
\label{tab6}
\end{center}
\end{table}

The results are reported in Table \ref{tab6}. Compared to the PACS dataset, the domain shift of OfficeHome is relatively smaller. Therefore, the  na\"ive DeepAll Baseline can get a good performance. Many DG approaches like CrossGrad\cite{shankar2018generalizing}, CCSA\cite{motiian2017unified}, and Jigen\cite{FabioMariaCarlucci2019DomainGB} can not improve much over the DeepAll. 
Thanks to the Fourier data augmentation, our approach gets a wonderful accuracy on the hardest domain clipart, which has a larger domain gap than others.
Finally, the top performance of average accuracy reveals the effectiveness of our approach.

\subsection{Evaluation on DG-Digits}
\begin{table}[htbp]
\caption{Model accuracy of leave-one-domain-out evaluation on Digits-DG. The best and second-best results are bolded and underlined, respectively.}
\begin{center}
\resizebox{\linewidth}{!}{
\begin{tabular}{l|cccc|c}
\hline
Methods          & MNIST            & MNIST-M        & SVHN          & SYN         & Avg.           \\ \hline
DeepAll          & 95.8          & 58.8          & 61.7 & 78.6          & 73.7          \\
CCSA\cite{motiian2017unified}            & 95.2          & 58.2         & 65.5          & 79.1          & 74.5          \\
MMD-AAE\cite{li2018domain}             & 96.5          & 58.4          & 65.0          & 78.4          & 74.6          \\
CrossGrad\cite{shankar2018generalizing}           & 96.7          & 61.1          & 65.3          & 80.2          & 75.8          \\
DDAIG\cite{zhou2020deep}           & 96.6          & 64.1          & 68.6          & 81.0          & 77.6          \\
Jigen\cite{FabioMariaCarlucci2019DomainGB}           & 96.5          & 61.4          & 63.7         & 74.0          & 73.9          \\
L2A-OT\cite{zhou2020learning}           & 96.7          & 63.9          & 68.6          & 83.2          & 78.1          \\
FACT\cite{xu2021fourier}             & \textbf{97.9}          & \underline{65.6}         & \underline{72.4}          & \underline{90.3} & \underline{81.5}          \\
\textbf{Ours}             & \underline{97.8}$\pm$0.1    & \textbf{66.8}$\pm$0.3 & \textbf{73.2}$\pm$0.5    & \textbf{90.7}$\pm$0.2 & \textbf{82.1} \\ \hline
\end{tabular}
}
\label{tab7}
\end{center}
\end{table}

\begin{table*}[htbp]
\begin{center}
\caption{Ablation studies on different choices of our strategy on the PACS dataset. \\The difference with our method is emphasized by red colors.}
\begin{tabular}{c|cccccccc|c}
\hline
\multirow{2}{*}{Pair relationship} & \multicolumn{2}{c|}{intra-domain} & \multicolumn{2}{c|}{intra-domain} & \multicolumn{2}{c|}{inter-domain} & \multicolumn{2}{c|}{inter-domain} & \multirow{3}{*}{Results} \\
                                   & \multicolumn{2}{c|}{intra-label}  & \multicolumn{2}{c|}{inter-label}  & \multicolumn{2}{c|}{intra-label}  & \multicolumn{2}{c|}{inter-label}  &                          \\
\cline{1-9}
mixup spectrums                   & Amplitude     & Phase            & Amplitude     & Phase            & Amplitude     & Phase            & Amplitude     & Phase            &                          \\
\hline
Ours                       &$\usym{2717}$  &$\usym{2713}$     &$\usym{2717}$  &$\usym{2713}$     &$\usym{2713}$  &$\usym{2713}$     &$\usym{2713}$  &$\usym{2717}$          & \textbf{88.55}  \\
Model A               &\textcolor{red}{$\usym{2713}$}  &$\usym{2713}$     &$\usym{2717}$  &$\usym{2713}$     &$\usym{2713}$  &$\usym{2713}$     &$\usym{2713}$  &$\usym{2717}$     & 88.07  \\ 
Model B               &$\usym{2717}$  &\textcolor{red}{$\usym{2717}$}     &$\usym{2717}$  &$\usym{2713}$     &$\usym{2713}$  &$\usym{2713}$     &$\usym{2713}$  &$\usym{2717}$     & 88.20  \\ 
Model C               &$\usym{2717}$  &$\usym{2713}$     &\textcolor{red}{$\usym{2713}$}  &$\usym{2713}$     &$\usym{2713}$  &$\usym{2713}$     &$\usym{2713}$  &$\usym{2717}$     & 87.87  \\ 
Model D               &$\usym{2717}$  &$\usym{2713}$     &$\usym{2717}$  &\textcolor{red}{$\usym{2717}$}     &$\usym{2713}$  &$\usym{2713}$     &$\usym{2713}$  &$\usym{2717}$     & 88.26  \\ 
Model E               &$\usym{2717}$  &$\usym{2713}$     &$\usym{2717}$  &$\usym{2713}$     &\textcolor{red}{$\usym{2717}$}  &$\usym{2713}$     &$\usym{2713}$  &$\usym{2717}$     & 88.32  \\ 
Model F               &$\usym{2717}$  &$\usym{2713}$     &$\usym{2717}$  &$\usym{2713}$     &$\usym{2713}$  &\textcolor{red}{$\usym{2717}$}     &$\usym{2713}$  &$\usym{2717}$     & 88.03  \\ 
Model G               &$\usym{2717}$  &$\usym{2713}$     &$\usym{2717}$  &$\usym{2713}$     &$\usym{2713}$  &$\usym{2713}$     &\textcolor{red}{$\usym{2717}$}  &$\usym{2717}$     & 87.88  \\ 
Model H               &$\usym{2717}$  &$\usym{2713}$     &$\usym{2717}$  &$\usym{2713}$     &$\usym{2713}$  &$\usym{2713}$     &$\usym{2713}$  &\textcolor{red}{$\usym{2713}$}     & 87.75  \\ 
\hline
\end{tabular} 
\label{tab5}
\end{center}
\end{table*}

We report the results in Table \ref{tab7}. In general, FACT\cite{xu2021fourier} and our method achieve the top-2 performance. Our approach gets the best on average since the markedly higher accuracy on the MNIST-M and SVHN domains. This again certifies the superiority of our approach.

\subsection{Visualization of the modified Fourier data augmentation}
In order to provide an intuitive illustration of SAM, we visualize the images augmented by different kinds of interpolation, as shown in Fig.\ref{fig4}. 
Note that in this visualization, the parameter $\lambda$ for the mixup, which should have been a random number obeying uniform distribution, is set to 0.5 for convenient observation. Therefore, two augmented images generated by mixup on both spectrums are the same ones.

 For inter-domain inter-label image pairs, Fig.\ref{fig4} shows that mixup on the phase leads to chaos of the augmented images, which is difficult for the model to learn valuable information, and the ablation study below is consistent with the conclusion.

\subsection{Ablation Studies}
To validate the effectiveness of our strategy, we design an ablation study as shown in Table \ref{tab5}. We pay attention to the model with one different choice during Fourier data augmentation compared with our approach. Every model is run three times independently on the PACS dataset, and the average accuracy of four domains is reported. 

In general, our approach achieves a better performance among all ablation models. 
Specifically, the results of Model A and Model C show that when dealing with intra-domain image pairs, doing mixup on the amplitude spectrum (which mainly contains style information) has no positive effect on the generalization ability. 
Besides, Model H certifies the correctness of removing the mixup of phase spectrums for inter-domain inter-label image pairs. However, for other situations where at least one property (domain and label) is the same, interpolation on the phase spectrums makes a positive difference, shown by Model B, Model D, and Model F.
Finally, the accuracy of Model E and Model G shows the great importance of doing mixup on style (amplitude) information for different domains to improve generalizability.

\section{Conclusion}
In this work, we propose a semantic-aware mixup (SAM) for domain generalization. The main idea is built upon the Fourier assumption that the phase spectrum represents semantic information and the amplitude spectrum contains style information. Accordingly, we propose that both the amplitude (style) and phase (semantic) information should be considered during data augmentation. Extensive experiments on the benchmarks demonstrate that our method can achieve state-of-the-art performance for domain generalization.




\section*{Acknowledgment}
This work was supported by NSFC No. 62222117 and the Fundamental Research Funds for the Central Universities under contract WK3490000005.

\bibliographystyle{unsrt}
\bibliography{ref}

\begin{thebibliography}{10}

\bibitem{AlexKrizhevsky2012ImageNetCW}
Alex Krizhevsky, Ilya Sutskever, and Geoffrey~E. Hinton.
\newblock Imagenet classification with deep convolutional neural networks.
\newblock {\em NeurIPS}, 2012.

\bibitem{JonathanTompson2014JointTO}
Jonathan Tompson, Arjun Jain, Yann LeCun, and Christoph Bregler.
\newblock Joint training of a convolutional network and a graphical model for
  human pose estimation.
\newblock {\em arXiv: CVPR}, 2014.

\bibitem{TomasMikolov2011StrategiesFT}
Tomas Mikolov, Anoop Deoras, Daniel Povey, Lukas Burget, and Jan Cernocky.
\newblock Strategies for training large scale neural network language models.
\newblock {\em IEEE Automatic Speech Recognition and Understanding Workshop},
  2011.

\bibitem{GeoffreyEHinton2012DeepNN}
Geoffrey~E. Hinton, Li~Deng, Dong Yu, George~E. Dahl, Abdelrahman Mohamed,
  Navdeep Jaitly, Andrew~W. Senior, Vincent Vanhoucke, Patrick Nguyen, Tara~N.
  Sainath, and Brian Kingsbury.
\newblock Deep neural networks for acoustic modeling in speech recognition: The
  shared views of four research groups.
\newblock {\em IEEE Signal Processing Magazine}, 2012.

\bibitem{IlyaSutskever2014SequenceTS}
Ilya Sutskever, Oriol Vinyals, and Quoc~V. Le.
\newblock Sequence to sequence learning with neural networks.
\newblock 2014.

\bibitem{RuhiSarikaya2014ApplicationOD}
Ruhi Sarikaya, Geoffrey~E. Hinton, and Anoop Deoras.
\newblock Application of deep belief networks for natural language
  understanding.
\newblock 2014.

\bibitem{IanGoodfellow2016DeepL}
Ian Goodfellow, Yoshua Bengio, and Aaron Courville.
\newblock Deep learning.
\newblock {\em MIT Press eBooks}, 2016.

\bibitem{YannLeCun2015DeepL}
Yann LeCun, Yoshua Bengio, and Geoffrey~E. Hinton.
\newblock Deep learning.
\newblock {\em Nature}, 2015.

\bibitem{KaimingHe2015DeepRL}
Kaiming He, Xiangyu Zhang, Shaoqing Ren, and Jian Sun.
\newblock Deep residual learning for image recognition.
\newblock pages 770--778, 2016.

\bibitem{DaLi2017DeeperBA}
Da~Li, Yongxin Yang, Yi-Zhe Song, and Timothy~M. Hospedales.
\newblock Deeper, broader and artier domain generalization.
\newblock {\em CVPR}, 2017.

\bibitem{RohanTaori2020MeasuringRT}
Rohan Taori, Achal Dave, Vaishaal Shankar, Nicholas Carlini, Benjamin Recht,
  and Ludwig Schmidt.
\newblock Measuring robustness to natural distribution shifts in image
  classification.
\newblock {\em NeurIPS}, 2020.

\bibitem{ShaiBenDavid2010ATO}
Shai Ben-David, John Blitzer, Koby Crammer, Alex Kulesza, Fernando Pereira, and
  Jennifer~Wortman Vaughan.
\newblock A theory of learning from different domains.
\newblock {\em Machine Learning}, 2010.

\bibitem{BenjaminRecht2019DoIC}
Benjamin Recht, Rebecca Roelofs, Ludwig Schmidt, and Vaishaal Shankar.
\newblock Do imagenet classifiers generalize to imagenet.
\newblock {\em arXiv: CVPR}, 2019.

\bibitem{JoseGMorenoTorres2012AUV}
Jose~G. Moreno-Torres, Troy Raeder, Roc{\'i}o Alaiz-Rodr{\'i}guez, Nitesh~V.
  Chawla, and Francisco Herrera.
\newblock A unifying view on dataset shift in classification.
\newblock {\em Pattern Recognition}, 2012.

\bibitem{wang2018deep}
Mei Wang and Weihong Deng.
\newblock Deep visual domain adaptation: A survey.
\newblock {\em Neurocomputing}, 312:135--153, 2018.

\bibitem{muandet2013domain}
Krikamol Muandet, David Balduzzi, and Bernhard Sch{\"o}lkopf.
\newblock Domain generalization via invariant feature representation.
\newblock In {\em ICML}, pages 10--18. PMLR, 2013.

\bibitem{fangout}
Zhen Fang, Yixuan Li, Jie Lu, Jiahua Dong, Bo~Han, and Feng Liu.
\newblock Is out-of-distribution detection learnable?
\newblock In {\em NeurIPS}.

\bibitem{li2018deep}
Ya~Li, Xinmei Tian, Mingming Gong, Yajing Liu, Tongliang Liu, Kun Zhang, and
  Dacheng Tao.
\newblock Deep domain generalization via conditional invariant adversarial
  networks.
\newblock In {\em ECCV}, pages 624--639, 2018.

\bibitem{li2018domain}
Haoliang Li, Sinno~Jialin Pan, Shiqi Wang, and Alex~C Kot.
\newblock Domain generalization with adversarial feature learning.
\newblock In {\em CVPR}, pages 5400--5409, 2018.

\bibitem{shao2019multi}
Rui Shao, Xiangyuan Lan, Jiawei Li, and Pong~C Yuen.
\newblock Multi-adversarial discriminative deep domain generalization for face
  presentation attack detection.
\newblock In {\em CVPR}, pages 10023--10031, 2019.

\bibitem{DaLi2017LearningTG}
Da~Li, Yongxin Yang, Yi-Zhe Song, and Timothy~M. Hospedales.
\newblock Learning to generalize: Meta-learning for domain generalization.
\newblock {\em arXiv: Learning}, 2017.

\bibitem{YogeshBalaji2018MetaRegTD}
Yogesh Balaji, Swami Sankaranarayanan, and Rama Chellappa.
\newblock Metareg: towards domain generalization using meta-regularization.
\newblock {\em NeurIPS}, 31, 2018.

\bibitem{li2019feature}
Yiying Li, Yongxin Yang, Wei Zhou, and Timothy Hospedales.
\newblock Feature-critic networks for heterogeneous domain generalization.
\newblock In {\em ICML}, pages 3915--3924. PMLR, 2019.

\bibitem{dou2019domain}
Qi~Dou, Daniel Coelho~de Castro, Konstantinos Kamnitsas, and Ben Glocker.
\newblock Domain generalization via model-agnostic learning of semantic
  features.
\newblock {\em NeurIPS}, 32, 2019.

\bibitem{FabioMariaCarlucci2019DomainGB}
Fabio~Maria Carlucci, Antonio D'Innocente, Silvia Bucci, Barbara Caputo, and
  Tatiana Tommasi.
\newblock Domain generalization by solving jigsaw puzzles.
\newblock {\em CVPR}, 2019.

\bibitem{volpi2018generalizing}
Riccardo Volpi, Hongseok Namkoong, Ozan Sener, John~C Duchi, Vittorio Murino,
  and Silvio Savarese.
\newblock Generalizing to unseen domains via adversarial data augmentation.
\newblock {\em NeurIPS}, 31, 2018.

\bibitem{shankar2018generalizing}
Shiv Shankar, Vihari Piratla, Soumen Chakrabarti, Siddhartha Chaudhuri, Preethi
  Jyothi, and Sunita Sarawagi.
\newblock Generalizing across domains via cross-gradient training.
\newblock {\em arXiv preprint arXiv:1804.10745}, 2018.

\bibitem{zhou2020deep}
Kaiyang Zhou, Yongxin Yang, Timothy Hospedales, and Tao Xiang.
\newblock Deep domain-adversarial image generation for domain generalisation.
\newblock In {\em AAAI}, volume~34, pages 13025--13032, 2020.

\bibitem{zhou2020learning}
Kaiyang Zhou, Yongxin Yang, Timothy Hospedales, and Tao Xiang.
\newblock Learning to generate novel domains for domain generalization.
\newblock In {\em ECCV}, pages 561--578. Springer, 2020.

\bibitem{xu2021fourier}
Qinwei Xu, Ruipeng Zhang, Ya~Zhang, Yanfeng Wang, and Qi~Tian.
\newblock A fourier-based framework for domain generalization.
\newblock In {\em CVPR}, pages 14383--14392, 2021.

\bibitem{AlanVOppenheim1979PhaseIS}
Alan~V. Oppenheim, Jae Lim, G.~Kopec, and S.~C. Pohlig.
\newblock Phase in speech and pictures.
\newblock {\em international conference on acoustics, speech, and signal
  processing}, 1979.

\bibitem{AlanVOppenheim1981TheIO}
Alan~V. Oppenheim and Jae Lim.
\newblock The importance of phase in signals.
\newblock {\em Proceedings of the IEEE}, 1981.

\bibitem{BruceCHansen2007StructuralSA}
Bruce~C. Hansen and Robert~F. Hess.
\newblock Structural sparseness and spatial phase alignment in natural scenes.
\newblock {\em Journal of The Optical Society of America A-optics Image Science
  and Vision}, 2007.

\bibitem{LeonNPiotrowski1982ADO}
Leon~N Piotrowski and F.~W. Campbell.
\newblock A demonstration of the visual importance and flexibility of
  spatial-frequency amplitude and phase.
\newblock {\em Perception}, 1982.

\bibitem{yang2020fda}
Yanchao Yang and Stefano Soatto.
\newblock Fda: Fourier domain adaptation for semantic segmentation.
\newblock In {\em CVPR}, pages 4085--4095, 2020.

\bibitem{HongyiZhang2017mixupBE}
Hongyi Zhang, Moustapha Cisse, Yann~N. Dauphin, and David Lopez-Paz.
\newblock mixup: Beyond empirical risk minimization.
\newblock {\em Learning}, 2017.

\bibitem{pang2018towards}
Tianyu Pang, Chao Du, Yinpeng Dong, and Jun Zhu.
\newblock Towards robust detection of adversarial examples.
\newblock {\em NeurIPS}, 31, 2018.

\bibitem{causal}
Yonggang Zhang, Mingming Gong, Tongliang Liu, Gang Niu, Xinmei Tian, Bo~Han,
  Bernhard Sch{\"o}lkopf, and Kun Zhang.
\newblock Causaladv: Adversarial robustness through the lens of causality.
\newblock In {\em ICLR}, 2022.

\bibitem{HemanthVenkateswara2017DeepHN}
Hemanth Venkateswara, Jose Eusebio, Shayok Chakraborty, and Sethuraman
  Panchanathan.
\newblock Deep hashing network for unsupervised domain adaptation.
\newblock {\em Springer International Publishing eBooks}, 2017.

\bibitem{zhou2022domain}
Kaiyang Zhou, Ziwei Liu, Yu~Qiao, Tao Xiang, and Chen~Change Loy.
\newblock Domain generalization: A survey.
\newblock {\em IEEE Transactions on Pattern Analysis and Machine Intelligence},
  2022.

\bibitem{ghifary2016scatter}
Muhammad Ghifary, David Balduzzi, W~Bastiaan Kleijn, and Mengjie Zhang.
\newblock Scatter component analysis: A unified framework for domain adaptation
  and domain generalization.
\newblock {\em IEEE transactions on pattern analysis and machine intelligence},
  39(7):1414--1430, 2016.

\bibitem{jia2020single}
Yunpei Jia, Jie Zhang, Shiguang Shan, and Xilin Chen.
\newblock Single-side domain generalization for face anti-spoofing.
\newblock In {\em CVPR}, pages 8484--8493, 2020.

\bibitem{li2018learning}
Da~Li, Yongxin Yang, Yi-Zhe Song, and Timothy Hospedales.
\newblock Learning to generalize: Meta-learning for domain generalization.
\newblock In {\em AAAI}, volume~32, 2018.

\bibitem{finn2017model}
Chelsea Finn, Pieter Abbeel, and Sergey Levine.
\newblock Model-agnostic meta-learning for fast adaptation of deep networks.
\newblock In {\em ICML}, pages 1126--1135. PMLR, 2017.

\bibitem{shi2020informative}
Baifeng Shi, Dinghuai Zhang, Qi~Dai, Zhanxing Zhu, Yadong Mu, and Jingdong
  Wang.
\newblock Informative dropout for robust representation learning: A shape-bias
  perspective.
\newblock In {\em ICML}, pages 8828--8839. PMLR, 2020.

\bibitem{geirhos2018imagenet}
Robert Geirhos, Patricia Rubisch, Claudio Michaelis, Matthias Bethge, Felix~A
  Wichmann, and Wieland Brendel.
\newblock Imagenet-trained cnns are biased towards texture; increasing shape
  bias improves accuracy and robustness.
\newblock {\em arXiv preprint arXiv:1811.12231}, 2018.

\bibitem{wang2020learning}
Shujun Wang, Lequan Yu, Caizi Li, Chi-Wing Fu, and Pheng-Ann Heng.
\newblock Learning from extrinsic and intrinsic supervisions for domain
  generalization.
\newblock In {\em ECCV}, pages 159--176. Springer, 2020.

\bibitem{yang2020phase}
Yanchao Yang, Dong Lao, Ganesh Sundaramoorthi, and Stefano Soatto.
\newblock Phase consistent ecological domain adaptation.
\newblock In {\em CVPR}, pages 9011--9020, 2020.

\bibitem{hoffman2018cycada}
Judy Hoffman, Eric Tzeng, Taesung Park, Jun-Yan Zhu, Phillip Isola, Kate
  Saenko, Alexei Efros, and Trevor Darrell.
\newblock Cycada: Cycle-consistent adversarial domain adaptation.
\newblock In {\em ICML}, pages 1989--1998. Pmlr, 2018.

\bibitem{laine2016temporal}
Samuli Laine and Timo Aila.
\newblock Temporal ensembling for semi-supervised learning.
\newblock {\em arXiv preprint arXiv:1610.02242}, 2016.

\bibitem{tarvainen2017mean}
Antti Tarvainen and Harri Valpola.
\newblock Mean teachers are better role models: Weight-averaged consistency
  targets improve semi-supervised deep learning results.
\newblock {\em NeurIPS}, 30, 2017.

\bibitem{miyato2018virtual}
Takeru Miyato, Shin-ichi Maeda, Masanori Koyama, and Shin Ishii.
\newblock Virtual adversarial training: a regularization method for supervised
  and semi-supervised learning.
\newblock {\em IEEE transactions on pattern analysis and machine intelligence},
  41(8):1979--1993, 2018.

\bibitem{park2018adversarial}
Sungrae Park, JunKeon Park, Su-Jin Shin, and Il-Chul Moon.
\newblock Adversarial dropout for supervised and semi-supervised learning.
\newblock In {\em AAAI}, volume~32, 2018.

\bibitem{verma2019interpolation}
Vikas Verma, Kenji Kawaguchi, Alex Lamb, Juho Kannala, Yoshua Bengio, and David
  Lopez-Paz.
\newblock Interpolation consistency training for semi-supervised learning.
\newblock {\em arXiv preprint arXiv:1903.03825}, 2019.

\bibitem{french2017self}
Geoffrey French, Michal Mackiewicz, and Mark Fisher.
\newblock Self-ensembling for visual domain adaptation.
\newblock {\em arXiv preprint arXiv:1706.05208}, 2017.

\bibitem{shu2018dirt}
Rui Shu, Hung~H Bui, Hirokazu Narui, and Stefano Ermon.
\newblock A dirt-t approach to unsupervised domain adaptation.
\newblock {\em arXiv preprint arXiv:1802.08735}, 2018.

\bibitem{wu2020dual}
Yuan Wu, Diana Inkpen, and Ahmed El-Roby.
\newblock Dual mixup regularized learning for adversarial domain adaptation.
\newblock In {\em ECCV}, pages 540--555. Springer, 2020.

\bibitem{angel1982fast}
ES~Angel.
\newblock Fast fourier transform and convolution algorithm.
\newblock {\em Proceedings of the IEEE}, 70(5):527--527, 1982.

\bibitem{he2020momentum}
Kaiming He, Haoqi Fan, Yuxin Wu, Saining Xie, and Ross Girshick.
\newblock Momentum contrast for unsupervised visual representation learning.
\newblock In {\em CVPR}, pages 9729--9738, 2020.

\bibitem{AnttiTarvainen2017MeanTA}
Antti Tarvainen and Harri Valpola.
\newblock Mean teachers are better role models: Weight-averaged consistency
  targets improve semi-supervised deep learning results.
\newblock {\em ICLR}, 2017.

\bibitem{YannLeCun1998GradientbasedLA}
Yann LeCun, L{\'e}on Bottou, Yoshua Bengio, and Patrick Haffner.
\newblock Gradient-based learning applied to document recognition.
\newblock {\em Proceedings of the IEEE}, 1998.

\bibitem{YaroslavGanin2015UnsupervisedDA}
Yaroslav Ganin and Victor Lempitsky.
\newblock Unsupervised domain adaptation by backpropagation.
\newblock {\em ICML}, 2015.

\bibitem{YuvalNetzer2011ReadingDI}
Yuval Netzer, Tao Wang, Adam Coates, Alessandro Bissacco, Bo~Wu, and Andrew~Y.
  Ng.
\newblock Reading digits in natural images with unsupervised feature learning.
\newblock 2011.

\bibitem{JiaDeng2009ImageNetAL}
Jia Deng, Wei Dong, Richard Socher, Li-Jia Li, Kai Li, and Li~Fei-Fei.
\newblock Imagenet: A large-scale hierarchical image database.
\newblock {\em CVPR}, 2009.

\bibitem{QiDou2019DomainGV}
Qi~Dou, Daniel~Coelho de~Castro, Konstantinos Kamnitsas, and Ben Glocker.
\newblock Domain generalization via model-agnostic learning of semantic
  features.
\newblock {\em NeurIPS}, 32, 2019.

\bibitem{ShujunWang2020LearningFE}
Shujun Wang, Lequan Yu, Caizi Li, Chi-Wing Fu, and Pheng-Ann Heng.
\newblock Learning from extrinsic and intrinsic supervisions for domain
  generalization.
\newblock pages 159--176, 2020.

\bibitem{ZeyiHuang2020SelfchallengingIC}
Zeyi Huang, Haohan Wang, Eric~P Xing, and Dong Huang.
\newblock Self-challenging improves cross-domain generalization.
\newblock pages 124--140, 2020.

\bibitem{motiian2017unified}
Saeid Motiian, Marco Piccirilli, Donald~A Adjeroh, and Gianfranco Doretto.
\newblock Unified deep supervised domain adaptation and generalization.
\newblock In {\em ICCV}, pages 5715--5725, 2017.

\end{thebibliography}

\end{document}